\documentclass[11pt]{article}
\usepackage[preprint]{acl}
\usepackage{times}
\usepackage{latexsym}
\usepackage[T1]{fontenc}
\usepackage[utf8]{inputenc}
\usepackage{microtype}
\usepackage{url}
\usepackage{inconsolata}
\usepackage{graphicx}
\usepackage{booktabs}
\usepackage{multirow}
\usepackage{tikz}
\usetikzlibrary{arrows.meta,positioning}
\usepackage[breakable]{tcolorbox}
\usepackage{fvextra}
\usepackage{hyperref}

\newcommand\blfootnote[1]{%
  \begingroup
  \renewcommand\thefootnote{}\footnote{#1}%
  \addtocounter{footnote}{-1}%
  \endgroup
}

\title{Finding the Right Tables and Columns: A Benchmark and Corpus-Adaptive Embeddings for SQL Schema Retrieval}

\author{
Qingcheng Zeng\textsuperscript{1,2},
Puxuan Yu\textsuperscript{2},
Aman Mehta\textsuperscript{2},
Fuheng Zhao\textsuperscript{2},
Rajhans Samdani\textsuperscript{2}\\
\textsuperscript{1}Northwestern University \quad
\textsuperscript{2}Snowflake Inc.\\[0.7em]
\normalfont\raisebox{-0.2em}{\includegraphics[height=1.15em]{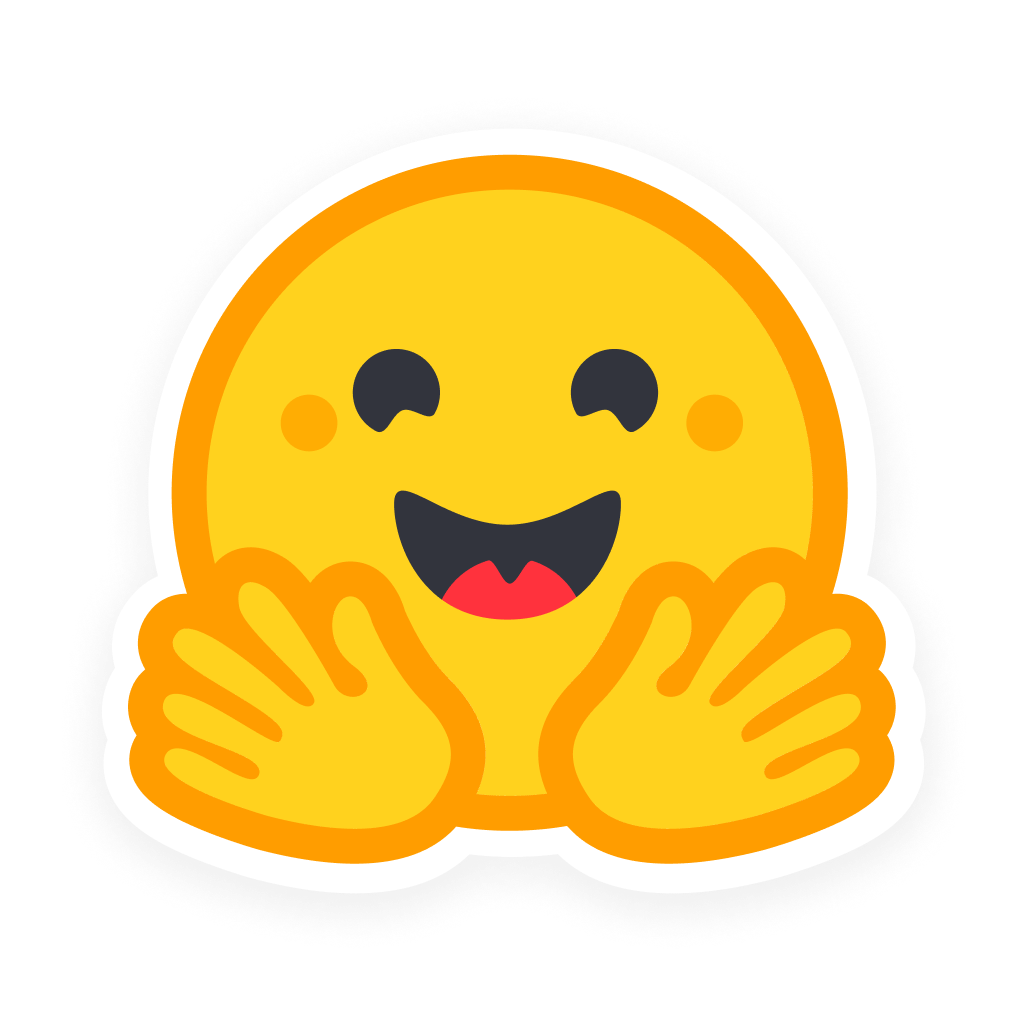}}\hspace{0.3em}%
\textbf{Dataset:}\hspace{0.4em}%
\href{https://huggingface.co/datasets/Snowflake/SQL-Schema-Retrieval}{\small\texttt{hf.co/datasets/Snowflake/SQL-Schema-Retrieval}}\\
}

\begin{document}
\maketitle
% The disclosure below is suppressed automatically in review mode and
% reappears when the [preprint] or [final] option is restored.
\makeatletter
\ifacl@anonymize\else
  \blfootnote{Work done during Qingcheng Zeng's internship at Snowflake Inc.}
\fi
\makeatother

\begin{abstract}
Retrieval in the SQL setting has largely been studied as the task of finding, within a large collection of SQL statements, the statement that answers a natural-language question. At scale, however, a more fundamental retrieval problem precedes generation: \emph{schema retrieval}, identifying the tables and columns a question requires in a database that may contain thousands of them, far more than fit in a model's context. We argue that this step warrants first-class evaluation. To this end, we recast five text-to-SQL datasets (Spider, BIRD, BEAVER, and two LiveSQLBench variants) as retrieval tasks at both table and column granularity, covering realistic and enterprise-scale schemas under two document representations, and we show that off-the-shelf text and code embedders transfer poorly to this setting. We then propose \emph{corpus-adaptive} fine-tuning: natural-language queries are synthesized directly from the target schema corpus, granularity-aware hard negatives are mined, and a 305M-parameter embedder is fine-tuned contrastively. This procedure raises average recall@10 from 60.4 to 75.6 (nDCG@10 from 51.9 to 68.0), making the 305M model the strongest retriever under one billion parameters and competitive with state-of-the-art embedders of 4--8B parameters, more than an order of magnitude larger. The same recipe improves an 8B state-of-the-art embedder from 77.8 to 78.4 recall@10, matching the best result on the benchmark and indicating that the adaptation is backbone-agnostic. Leave-one-corpus-out experiments and a leakage audit show that these gains reflect a transferable schema-retrieval ability rather than memorization of the evaluation data. Our results establish schema linking as a standalone retrieval task and lightweight, label-free corpus adaptation as a practical route to deploying it at enterprise scale.
\end{abstract}

\section{Introduction}

Information retrieval over code has emerged as a distinct evaluation target, with benchmarks such as CoIR \citep{li-etal-2025-coir} and CoQuIR \citep{geng-etal-2026-coquir} measuring how well embedding models retrieve code given a natural-language intent. SQL increasingly falls within this scope: text-to-SQL, long studied as a generation task on benchmarks such as WikiSQL \citep{zhong2017seq2sqlgeneratingstructuredqueries}, Spider \citep{yu-etal-2018-spider}, and BIRD \citep{3666122.3667957}, is now also framed as retrieval, in which a question is matched against a collection of SQL statements or code snippets. One retrieval step specific to databases, however, remains largely absent from these benchmarks: given a question and a database, retrieving the tables and columns that the question refers to.

This \emph{schema retrieval} step is indispensable at scale. Enterprise data warehouses routinely contain hundreds to thousands of tables and tens of thousands of columns \citep{chen2026beaverenterprisebenchmarktexttosql}, far more than fit in a model's context window or than a downstream component can reasonably process. Selecting the relevant schema elements first, a form of schema linking \citep{10.1609/aaai.v37i11.26535}, is therefore a prerequisite for database tasks over large schemas, and it constitutes a natural retrieval problem: ranking schema documents by their relevance to a query. Nevertheless, schema retrieval is rarely evaluated as a retrieval task in its own right; existing code- and SQL-retrieval benchmarks target whole programs or queries rather than the schema elements underlying them.

In this work, we establish schema retrieval as a first-class task. Our contributions are as follows:
\begin{itemize}
\itemsep0.1em
\item A \textbf{benchmark} that casts schema linking as document retrieval at both table and column granularity, derived from five text-to-SQL datasets spanning academic (Spider), realistic (BIRD), enterprise (BEAVER), and large live-database (LiveSQLBench) settings, under two document representations (\S\ref{sec:benchmark}).
\item An \textbf{evaluation} of general-purpose, code-specialized, and instruction-tuned embedders spanning 0.1B to 8B parameters, showing that even strong off-the-shelf models transfer poorly to schema retrieval (\S\ref{sec:results}).
\item \textbf{Corpus-adaptive fine-tuning} (\S\ref{sec:method}): given only the target schema corpus, we synthesize grounded queries and mine \emph{granularity-aware} hard negatives, raising a 305M-parameter embedder from 60.4 to 75.6 recall@10, the best result under one billion parameters and within three points of 4--8B models; the same recipe also improves an 8B state-of-the-art embedder (\S\ref{sec:results}).
\item \textbf{Validation}: leave-one-corpus-out experiments and a query/label leakage audit (\S\ref{sec:results}, Appendices~\ref{app:leakage}--\ref{app:loo}) demonstrate that the improvements stem from generalization rather than memorization.
\end{itemize}

\section{Related Work}

\paragraph{Text-to-SQL and schema linking.}
Cross-domain benchmarks \citep{yu-etal-2018-spider,3666122.3667957}, together with recent enterprise \citep{chen2026beaverenterprisebenchmarktexttosql} and large-database \citep{livesqlbench2025} settings, have driven rapid progress on text-to-SQL, but evaluation has centered on end-to-end execution accuracy. Schema linking, identifying the schema elements that a query requires, is a well-studied sub-problem \citep{10.1609/aaai.v37i11.26535,3666122.3667699,10.14778/3641204.3641221}; as schemas grow beyond what fits in context, it is increasingly implemented as a retrieval step whose output is passed to an LLM, which motivates evaluating it directly.

\paragraph{Dense retrieval and embedding models.}
Dual-encoder retrievers \citep{reimers-gurevych-2019-sentence,karpukhin-etal-2020-dense} and contrastively pre-trained text embedders \citep{izacard2021contriever,wang2024textembeddingsweaklysupervisedcontrastive,merrick2024arcticembedscalableefficientaccurate,10.1145/3626772.3657878} dominate first-stage retrieval, with late-interaction \citep{10.1145/3397271.3401075} and code-specialized \citep{liu2025codexembedgeneralistembeddingmodel,sureshcornstack} variants developed for structured inputs. Standard text- and code-retrieval benchmarks \citep{thakur2021beir,muennighoff-etal-2023-mteb,li-etal-2025-coir,geng-etal-2026-coquir} evaluate zero-shot transfer but do not cover schema retrieval, a setting in which we find that even strong general-purpose and code embedders underperform.

\paragraph{Synthetic data and domain adaptation.}
Generating queries from documents to train retrievers has a long lineage, spanning document expansion \citep{nogueira2019documentexpansionqueryprediction}, LLM-based query generation \citep{10.1145/3477495.3531863,dai2022promptagatorfewshotdenseretrieval}, and generative pseudo-labeling for domain adaptation \citep{wang-etal-2022-gpl}; hard-negative mining is central to the quality of the resulting training data \citep{xiong2020approximatenearestneighbornegative,qu-etal-2021-rocketqa}. Our method belongs to this lineage but is tailored to schema retrieval: it is \emph{granularity-aware}, jointly handling table- and column-level documents, and it adapts the embedder to the target corpus.

\section{The Schema-Retrieval Benchmark}
\label{sec:benchmark}

\paragraph{Task.}
Each database defines a corpus $D$ of schema documents. At \emph{table} granularity, each document describes one table; at \emph{column} granularity, each document describes one column together with its table context. Given a natural-language query $q$, the relevant set $R(q)\subseteq D$ comprises the schema elements referenced by the gold SQL; we obtain $R(q)$ by parsing table and column references from the ground-truth SQL. Systems rank the documents in $D$ for each query. We adopt recall@10 as the primary metric, since downstream generation requires the needed schema elements to appear among the top-ranked candidates, and additionally report nDCG@10. Retrieval is performed \emph{per database group}, mirroring deployment, in which a system retrieves within the schema of the database at hand.

\paragraph{Sources and representations.}
We derive the benchmark from five datasets (Table~\ref{tab:bench}): Spider \citep{yu-etal-2018-spider} (cross-domain, simple schemas), BIRD \citep{3666122.3667957} (realistic schemas with value semantics), BEAVER \citep{chen2026beaverenterprisebenchmarktexttosql} (private enterprise warehouses), and the base and large variants of LiveSQLBench \citep{livesqlbench2025} (large, obfuscated, real-world schemas). For BEAVER, BIRD, and Spider we include two document representations of each table: a \emph{schema-metadata} form (DDL with keys plus sample rows) and a \emph{value-only} form (sample rows alone).

\begin{table}[t]
\centering
\footnotesize
\setlength{\tabcolsep}{4pt}
\begin{tabular}{lrrrr}
\toprule
Dataset & \#DB & \#Tab & \#Col & \#Q \\
\midrule
Spider              & 40 & 180 & 785    & 2{,}147 \\
BIRD                & 11 & 75  & 798    & 1{,}534 \\
BEAVER              & 6  & 463 & 4{,}238  & 209 \\
LiveSQLBench        & 22 & 244 & 1{,}942  & 410 \\
LiveSQLBench-Large  & 18 & 971 & 17{,}708 & 332 \\
\bottomrule
\end{tabular}
\caption{Benchmark statistics. \#Q is the number of evaluation queries; gold sets average 1.6--4.4 tables (table level) and up to 6.5 columns (column level) per query.}
\label{tab:bench}
\end{table}

\section{Corpus-Adaptive Fine-Tuning}
\label{sec:method}

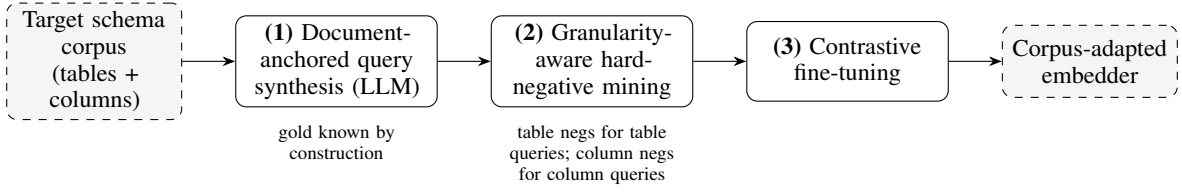
\begin{figure*}[t]
\centering
\footnotesize
\begin{tikzpicture}[
  box/.style={draw, rounded corners, align=center, text width=2.45cm, minimum height=1.05cm, inner sep=3pt},
  data/.style={draw, dashed, rounded corners, align=center, text width=2.1cm, minimum height=0.95cm, inner sep=3pt, fill=black!4},
  >={Stealth[]}, node distance=0.55cm and 0.7cm]
\node[data] (corpus) {Target schema corpus\\(tables + columns)};
\node[box, right=of corpus] (gen) {\textbf{(1)} Document-anchored query synthesis (LLM)};
\node[box, right=of gen] (mine) {\textbf{(2)} Granularity-aware hard-negative mining};
\node[box, right=of mine] (ft) {\textbf{(3)} Contrastive fine-tuning};
\node[data, right=of ft] (model) {Corpus-adapted embedder};
\draw[->] (corpus) -- (gen);
\draw[->] (gen) -- (mine);
\draw[->] (mine) -- (ft);
\draw[->] (ft) -- (model);
\node[below=0.12cm of gen, text width=2.6cm, align=center, font=\scriptsize] {gold known by construction};
\node[below=0.12cm of mine, text width=2.6cm, align=center, font=\scriptsize] {table negs for table queries; column negs for column queries};
\end{tikzpicture}
\caption{Corpus-adaptive fine-tuning. Queries are synthesized over the target schema corpus with gold sets fixed by construction; hard negatives are mined within each (source, granularity) pool; the embedder is fine-tuned contrastively.}
\label{fig:pipeline}
\end{figure*}

In in-domain retrieval, the corpus of schema documents is available at indexing time. We exploit this standard assumption to adapt an embedder to a target corpus \emph{without any labeled queries}, in three steps (Figure~\ref{fig:pipeline}).

\paragraph{(1) Document-anchored query synthesis.}
Prompting an LLM to write queries for an entire schema yields uneven coverage and biases generation away from abbreviated identifiers. We therefore \emph{anchor} generation on target documents: we sample schema elements (individual columns, coherent column groups, and foreign-key-connected table sets) and prompt an instruction-tuned LLM to write a natural-language query answerable by exactly those elements, without naming them literally. Because the target elements are fixed in advance, the gold set is known by construction, with no SQL parsing required, and coverage of every table and column, and of both granularities, is guaranteed. We additionally generate queries for both document representations. The full prompt templates are given in Appendix~\ref{app:prompts}.

\paragraph{(2) Granularity-aware hard negatives.}
We embed all queries and documents and, for each query, retrieve the nearest non-gold documents as hard negatives \citep{xiong2020approximatenearestneighbornegative}. Crucially, negatives are mined \emph{within the same (source, granularity) pool} against which the query is evaluated: table-level queries draw table-document negatives, and column-level queries draw column-document negatives, excluding the gold columns' parent tables and sibling columns. This design avoids a failure mode that arises when negatives are mixed across granularities, in which table documents become negatives for column queries and vice versa, degrading retrieval at the opposite granularity.

\paragraph{(3) Contrastive fine-tuning.}
We fine-tune Arctic-Embed-M \citep{merrick2024arcticembedscalableefficientaccurate} (305M parameters) with a standard contrastive objective, using one positive and six hard negatives per query, and mix the synthesized in-domain data with existing schema-linking pairs to preserve general retrieval ability. Implementation details are given in Appendix~\ref{app:impl}.

\begin{table*}[t]
\centering
\footnotesize
\setlength{\tabcolsep}{3.5pt}
\begin{tabular}{l c ccc ccc ccc c}
\toprule
 & & \multicolumn{3}{c}{Value-only} & \multicolumn{3}{c}{Schema-metadata} & \multicolumn{3}{c}{LiveSQLBench} & \\
\cmidrule(lr){3-5}\cmidrule(lr){6-8}\cmidrule(lr){9-11}
Model & Size & BEA & BIR & SPI & BEA & BIR & SPI & Base & Lg.-T & Lg.-C & Avg \\
\midrule
BM25 & -- & 30.0 & 66.1 & 56.5 & 37.6 & 72.3 & 70.5 & 59.2 & 31.3 & 5.5 & 47.7 \\
Arctic-Embed-M & 0.3B & 12.5 & 75.3 & 67.3 & 42.2 & 85.3 & 95.0 & 86.6 & 59.1 & 20.5 & 60.4 \\
CodeRankEmbed & 0.1B & 39.4 & 67.5 & 92.9 & 43.9 & 71.0 & 95.9 & 58.5 & 50.3 & 21.7 & 60.1 \\
Arctic-Embed-L & 0.6B & 27.5 & 88.7 & 84.5 & 48.9 & 94.2 & 94.7 & 83.2 & 62.3 & 27.7 & 68.0 \\
E5-large & 0.3B & 32.2 & 93.6 & 88.6 & 43.0 & 97.3 & 93.2 & 74.5 & 57.6 & 20.9 & 66.8 \\
CodeXEmbed-400M & 0.4B & 42.8 & 91.5 & 86.5 & 62.9 & 97.5 & 96.3 & 87.1 & 62.9 & 18.0 & 71.7 \\
Qwen3-Embedding-0.6B & 0.6B & 43.6 & 94.4 & 92.6 & 52.0 & 96.2 & 97.1 & 87.0 & 67.5 & 24.4 & 72.8 \\
CodeXEmbed-2B & 2B & 28.7 & 88.7 & 88.8 & 36.5 & 90.9 & 95.3 & 86.8 & 59.1 & 17.0 & 65.7 \\
Qwen3-Embedding-4B & 4B & 49.3 & 96.8 & 95.0 & \textbf{65.8} & 97.5 & 97.9 & 90.6 & \textbf{75.8} & \textbf{37.0} & \textbf{78.4} \\
Qwen3-Embedding-8B & 8B & 52.1 & 95.4 & 94.4 & 63.9 & 98.5 & \textbf{98.5} & \textbf{91.7} & 69.5 & 36.2 & 77.8 \\
\midrule
\textbf{Corpus-adaptive Arctic-M (ours)} & 0.3B & 48.4 & 97.5 & 91.5 & 55.4 & 98.2 & 95.3 & 87.0 & 74.1 & 33.5 & 75.6 \\
\textbf{Corpus-adaptive Qwen3-8B (ours)} & 8B & \textbf{56.7} & \textbf{98.7} & \textbf{96.1} & 62.8 & \textbf{99.1} & 97.4 & 86.5 & 72.5 & 35.6 & 78.4 \\
\bottomrule
\end{tabular}
\caption{Schema-retrieval \textbf{recall@10} ($\times100$). BEA/BIR/SPI are BEAVER/BIRD/Spider table retrieval under value-only and schema-metadata document representations; Base/Lg.-T/Lg.-C are LiveSQLBench (base table), LiveSQLBench-Large (table) and (column). Best per column in \textbf{bold}. Our corpus-adaptive Arctic-M (0.3B) is the strongest model under 1B parameters; applying the same recipe to Qwen3-Embedding-8B matches the best overall recall@10 (78.4) and attains the best nDCG@10 (Table~\ref{tab:ndcg}).}
\label{tab:recall}
\end{table*}

\begin{table*}[t]
\centering
\footnotesize
\setlength{\tabcolsep}{3.5pt}
\begin{tabular}{l c ccc ccc ccc c}
\toprule
 & & \multicolumn{3}{c}{Value-only} & \multicolumn{3}{c}{Schema-metadata} & \multicolumn{3}{c}{LiveSQLBench} & \\
\cmidrule(lr){3-5}\cmidrule(lr){6-8}\cmidrule(lr){9-11}
Model & Size & BEA & BIR & SPI & BEA & BIR & SPI & Base & Lg.-T & Lg.-C & Avg \\
\midrule
BM25 & -- & 25.8 & 38.6 & 36.3 & 32.0 & 42.9 & 49.1 & 49.3 & 21.9 & 5.9 & 33.5 \\
Arctic-Embed-M & 0.3B & 11.2 & 62.8 & 53.8 & 39.8 & 74.2 & 84.0 & 74.6 & 46.6 & 20.5 & 51.9 \\
CodeRankEmbed & 0.1B & 35.6 & 60.3 & 82.7 & 39.4 & 64.1 & 85.5 & 49.6 & 42.2 & 21.0 & 53.4 \\
Arctic-Embed-L & 0.6B & 24.5 & 75.0 & 71.0 & 41.3 & 81.4 & 84.8 & 73.8 & 51.6 & 28.7 & 59.1 \\
E5-large & 0.3B & 25.5 & 80.6 & 75.3 & 36.2 & 84.9 & 81.2 & 59.4 & 46.4 & 21.3 & 56.8 \\
CodeXEmbed-400M & 0.4B & 36.6 & 80.5 & 74.5 & 52.3 & 88.3 & 86.8 & 76.4 & 49.2 & 17.1 & 62.4 \\
Qwen3-Embedding-0.6B & 0.6B & 37.1 & 82.6 & 81.6 & 42.5 & 84.1 & 88.9 & 76.7 & 55.5 & 22.7 & 63.5 \\
CodeXEmbed-2B & 2B & 24.8 & 74.6 & 76.2 & 33.1 & 78.0 & 84.2 & 76.0 & 44.8 & 16.1 & 56.4 \\
Qwen3-Embedding-4B & 4B & 43.3 & 89.0 & 85.2 & \textbf{58.6} & 90.0 & 90.7 & 80.9 & \textbf{65.4} & 35.4 & 70.9 \\
Qwen3-Embedding-8B & 8B & 43.6 & 86.0 & 83.8 & 57.6 & 91.0 & \textbf{91.1} & \textbf{82.5} & 56.0 & 35.3 & 69.7 \\
\midrule
\textbf{Corpus-adaptive Arctic-M (ours)} & 0.3B & 42.5 & 88.3 & 82.6 & 48.3 & 90.8 & 86.6 & 76.9 & 62.0 & 33.7 & 68.0 \\
\textbf{Corpus-adaptive Qwen3-8B (ours)} & 8B & \textbf{50.6} & \textbf{92.7} & \textbf{87.9} & 54.7 & \textbf{93.6} & 90.5 & 77.0 & 62.0 & \textbf{36.3} & \textbf{71.7} \\
\bottomrule
\end{tabular}
\caption{Schema-retrieval \textbf{nDCG@10} ($\times100$); columns as in Table~\ref{tab:recall}. Corpus-adaptive fine-tuning of Qwen3-Embedding-8B attains the best average (71.7), improving over the 8B base (69.7).}
\label{tab:ndcg}
\end{table*}

\section{Experiments and Results}
\label{sec:results}

\paragraph{Setup.}
We compare general-purpose \citep{merrick2024arcticembedscalableefficientaccurate,wang2024textembeddingsweaklysupervisedcontrastive}, code-specialized \citep{sureshcornstack,liu2025codexembedgeneralistembeddingmodel}, and instruction-tuned \citep{zhang2025qwen3embeddingadvancingtext} embedders ranging from 0.1B to 8B parameters, together with a BM25 lexical baseline, against our corpus-adaptive models (Tables~\ref{tab:recall} and~\ref{tab:ndcg}). We report recall@10 and nDCG@10.

\paragraph{Schema retrieval is challenging even for strong embedders.}
Among off-the-shelf models, only the 4B and 8B Qwen3-Embedding models exceed 75 recall@10 (Tables~\ref{tab:recall}--\ref{tab:ndcg}); smaller general-purpose and code embedders trail substantially. The BM25 lexical baseline is weakest overall (47.7 recall@10) and collapses to 5.5 on LiveSQLBench-Large column retrieval, where abbreviated identifiers share almost no surface forms with natural-language queries. Scale alone is not sufficient even within a model family: the 2B CodeXEmbed (65.7 recall@10) underperforms its 400M sibling (71.7), and the 8B Qwen3-Embedding trails the 4B model on the largest schemas (LiveSQLBench-Large). The value-only representation, which exposes only sample rows, is consistently the hardest setting, with most models below 45 recall@10 on BEAVER.

\paragraph{Corpus-adaptive fine-tuning is parameter-efficient.}
Our recipe raises the 305M Arctic-Embed-M from 60.4 to 75.6 recall@10 (51.9 to 68.0 nDCG@10), making it the strongest retriever under one billion parameters and competitive with the 4B and 8B Qwen3-Embedding models (78.4 and 77.8 recall@10, respectively) despite being more than an order of magnitude smaller; it surpasses the 2B CodeXEmbed by 9.9 recall@10 points. Gains are largest where schemas are hardest: on LiveSQLBench-Large, table retrieval improves from 59.1 to 74.1 recall@10 and column retrieval from 20.5 to 33.5. Training on both document representations sharply improves value-only retrieval (e.g., BIRD: $75.3\rightarrow97.5$ recall@10), confirming that representation coverage matters.

\paragraph{Corpus adaptation also improves a strong backbone.}
The recipe is not specific to a small encoder. Applying the identical pipeline, data, and objective to Qwen3-Embedding-8B (Appendix~\ref{app:impl}) raises its average recall@10 from 77.8 to 78.4, matching the strongest model on the benchmark (Qwen3-Embedding-4B), and raises nDCG@10 from 69.7 to 71.7, the highest of any model we evaluate. As with the 305M backbone, the recall gains are largest on the hardest, value-only settings (BEAVER $52.1\rightarrow56.7$, BIRD $95.4\rightarrow98.7$) and on LiveSQLBench-Large tables ($69.5\rightarrow72.5$), partly offset by a decline on LiveSQLBench base ($91.7\rightarrow86.5$). That adaptation improves even a near-state-of-the-art 8B embedder suggests that it complements scale rather than substitutes for it: the headroom our benchmark exposes is not closed simply by adopting a larger off-the-shelf model.

\paragraph{Ablation.}
We ablate the two ingredients that distinguish our recipe from naive document-anchored synthesis: table-level positives and granularity-aware negatives. Removing both (i.e., column-only synthesis with negatives mined across granularities) collapses table-level retrieval from 80.9 to 52.4 recall@10, below the base model's 65.4, while column-level retrieval declines moderately ($33.5\rightarrow29.2$) yet remains well above the base model's 20.5 (Table~\ref{tab:ablation}). Jointly modeling both granularities and isolating negatives by granularity is thus what preserves table-level retrieval.

\begin{table}[t]
\centering
\footnotesize
\setlength{\tabcolsep}{4pt}
\begin{tabular}{lccc}
\toprule
Variant & Table & Column & Avg \\
\midrule
Corpus-adaptive (full)             & \textbf{80.9} & \textbf{33.5} & \textbf{75.6} \\
\;\;$-$ table positives \& gran. negs.  & 52.4 & 29.2 & 49.8 \\
\midrule
Base Arctic-Embed-M (ref.)         & 65.4 & 20.5 & 60.4 \\
\bottomrule
\end{tabular}
\caption{Ablation (recall@10, $\times100$). \emph{Table} averages the eight table-level settings; \emph{Column} is LiveSQLBench-Large column retrieval. Dropping table positives and granularity-aware negatives pushes table-level retrieval below the base model.}
\label{tab:ablation}
\end{table}

\paragraph{The gains reflect generalization, not memorization.}
In leave-one-corpus-out experiments (Appendix~\ref{app:loo}), removing a dataset entirely from synthesis retains most of its gain: a model that never sees BEAVER still attains 44.3/57.7 (value-only/schema-metadata) recall@10, compared with 12.5/42.2 for the base model, and a model that never sees LiveSQLBench-Large retains 69.3 (table) and 32.0 (column). A query-overlap audit (Appendix~\ref{app:leakage}) finds that no synthetic query reconstructs an evaluation query with matching gold labels in five of six settings, and that only 2 of 2{,}147 do so in the sixth.

\paragraph{Error analysis: ordering vs.\ representation.}
We examine \emph{where} gold documents fall in the ranking (Table~\ref{tab:rankbuckets}). For table retrieval, every gold table lies within the top 1000 for all four models, and most misses sit just outside the top 10 (ranks 11--100): the relevant tables are present in the embedding neighborhood, and the errors are errors of ordering. Column retrieval is harder and is partly a failure of \emph{representation}: some gold columns never enter the top 1000 at all. This failure diminishes with scale, from 15.7\% of gold columns for the 0.3B base model to 5.5\% for the 8B base model. Corpus adaptation closes most of that gap for the small model (15.7\%$\rightarrow$8.8\% unretrieved, approaching the 8B base) while also lifting its top-10 recall. For the 8B backbone, adaptation improves top-10 ordering ($27.7\rightarrow29.4$) but increases the unretrieved fraction ($5.5\%\rightarrow11.2\%$), suggesting that adapting an already-strong encoder trades some representational coverage for sharper ordering near the top. Column-level headroom is therefore largely a matter of surfacing identifiers that the embedder cannot represent, a failure that scale alleviates and that corpus adaptation addresses at a fraction of the model size.

\begin{table}[t]
\centering
\footnotesize
\setlength{\tabcolsep}{2pt}
\begin{tabular}{@{}ll cccc@{}}
\toprule
Level & Model & $\le$10 & 11--100 & 101--1000 & $>$1000 \\
\midrule
\multirow{4}{*}{Table}  & Arctic-M (base) & 73.1 & 20.8 & 6.1 & 0.0 \\
                        & Ours (0.3B)     & 86.4 & 12.1 & 1.5 & 0.0 \\
                        & Qwen3-8B (base) & 88.6 & 9.5 & 1.9 & 0.0 \\
                        & Ours (8B)       & 89.5 & 9.3 & 1.2 & 0.0 \\
\midrule
\multirow{4}{*}{Column} & Arctic-M (base) & 15.8 & 24.5 & 44.1 & 15.7 \\
                        & Ours (0.3B)     & 28.2 & 34.0 & 29.0 & 8.8 \\
                        & Qwen3-8B (base) & 27.7 & 37.2 & 29.6 & 5.5 \\
                        & Ours (8B)       & 29.4 & 31.0 & 28.4 & 11.2 \\
\bottomrule
\end{tabular}
\caption{Rank bucket of gold documents (\% of all gold). Table-level gold always falls within the top 1000, so table-level errors are errors of ordering. For columns, the never-retrieved ($>$1000) fraction shrinks with scale (15.7\% for the 0.3B base model, 5.5\% for the 8B base model); adapting the 0.3B model cuts it to 8.8\%, approaching the 8B base, whereas adapting the 8B model improves top-10 recall at the cost of a larger unretrieved fraction (11.2\%).}
\label{tab:rankbuckets}
\end{table}

\paragraph{Abbreviated column names drive difficulty.}
Most gold columns are not named in natural language: 71\% of gold-column names are abbreviations or codes (e.g., \texttt{mtbfh}, \texttt{coordopsref}, \texttt{snopr\_cd}) rather than dictionary words. Table~\ref{tab:abbrev} reports column recall@10 by name type. All models degrade as names become less word-like, but the corpus-adapted models close much of this gap: they achieve the best results on the code bucket (31.4 for the 0.3B model and 32.3 for the 8B model), exceeding \emph{every} off-the-shelf model there, including the 4B and 8B Qwen3-Embedding models (both 26.2). Larger general-purpose models retain an advantage on word-like names (Qwen3-Embedding-4B: 34.2), but corpus adaptation yields its largest improvements precisely on the abbreviated identifiers that dominate real schemas, and it does so at both backbone scales.

\begin{table}[t]
\centering
\footnotesize
\setlength{\tabcolsep}{4pt}
\begin{tabular}{lccc}
\toprule
Model & Word & Mixed & Code \\
 & ($n{=}614$) & ($n{=}634$) & ($n{=}909$) \\
\midrule
Arctic-M (base, 0.3B)  & 17.8 & 15.5 & 14.6 \\
Qwen3-Embedding-4B     & \textbf{34.2} & \textbf{26.8} & 26.2 \\
Qwen3-Embedding-8B     & 31.1 & 26.5 & 26.2 \\
Ours (0.3B)            & 28.3 & 23.7 & 31.4 \\
Ours (8B)              & 28.0 & 26.7 & \textbf{32.3} \\
\bottomrule
\end{tabular}
\caption{Column recall@10 by gold-column name type. Most gold columns (71\%) are abbreviations or codes. The corpus-adapted models gain most on code-like names, exceeding both the 4B and 8B Qwen3-Embedding there; the 0.3B adapted model does so despite being more than an order of magnitude smaller.}
\label{tab:abbrev}
\end{table}

\paragraph{Difficulty scales with query breadth.}
Recall declines steeply as a query requires more schema elements. For table retrieval, our model's recall@10 falls from 97.3 on single-table queries to 38.0 on queries requiring six or more tables (compared with a decline from 82.0 to 32.7 for the base model); for column retrieval, it falls from 50.0 to 25.9 (40.9 to 13.4 for the base model). Adaptation helps across the entire range, nearly doubling recall on the hardest queries requiring six or more columns ($13.4\rightarrow25.9$), but queries that touch many tables or columns remain the most challenging setting. Adaptation most often recovers columns whose names share no token with the query: for a question about plants ``losing more than a quarter of their potential power and being offline more than one day out of every twenty,'' the base model ranks \texttt{mtbfh} (mean time between failures) 11th, whereas our model ranks it 8th, with similar recoveries for \texttt{coordopsref} ($11\rightarrow1$), \texttt{AnonLevel} ($12\rightarrow2$), and \texttt{stockturnrate} ($11\rightarrow6$).

\paragraph{Discussion.}
Our findings have three implications. First, schema retrieval is not solved by scale: even multi-billion-parameter embedders leave substantial headroom on enterprise and large-database settings, warranting its treatment as a first-class task rather than as a byproduct of text-to-SQL training. Second, corpus adaptation is an inexpensive, label-free, backbone-agnostic complement to scale: it makes a 305M model competitive with 4--8B embedders, improves even a near-state-of-the-art 8B embedder, and, because it requires only the schema corpus, is directly deployable in the retrieve-then-generate pipelines that motivate this work \citep{3495724.3496517,3666122.3667699}. Third, because the recipe requires only a document corpus and an LLM, it should extend beyond SQL to other forms of structured-resource retrieval, such as selecting APIs, functions, or knowledge-base entries for tool-augmented LLM agents.

\section*{Limitations}

Our study is limited to English schemas and questions. Corpus-adaptive fine-tuning assumes access to the target schema corpus at adaptation time; this assumption holds for indexed retrieval but does not address cold-start databases whose schemas are unavailable. Gold relevance is derived by parsing references from ground-truth SQL, which can miss columns that are used implicitly. Synthetic queries are produced by an LLM and may not fully match the stylistic distribution of questions posed by human analysts, particularly terse ones. Finally, we validate the recipe on two backbones, a 305M encoder and an 8B decoder; whether it extends to structured retrieval beyond SQL remains an open question.

\bibliography{custom}

\appendix

\section{Implementation Details}
\label{app:impl}
Synthetic queries are generated with a proprietary instruction-tuned LLM served via Snowflake Cortex; the recipe does not depend on a specific generator, and we leave a systematic comparison of query-synthesis models to future work. Hard negatives are mined with a long-context embedding model (\texttt{voyage-multilingual-2}) using exact nearest-neighbor search within each (source, granularity) pool; we skip the top-1 neighbor as a guard against false negatives and ensure that each query receives at least one lexically overlapping negative. We fine-tune Arctic-Embed-M with a contrastive objective, using one positive and six hard negatives per query, and mix the in-domain synthetic data with schema-linking pairs derived from the public text-to-SQL training splits of Spider and BIRD \citep{yu-etal-2018-spider,3666122.3667957}. Training uses AdamW with a learning rate of $3\times10^{-5}$ and a batch size of 32, for one epoch on 8~NVIDIA H200 GPUs. We apply the same pipeline and contrastive objective to Qwen3-Embedding-8B, changing only backbone-specific settings: last-token pooling, left padding, and a reduced learning rate of $5\times10^{-6}$.

\section{Leakage Audit}
\label{app:leakage}
For every evaluation query, we retrieve its most similar synthetic training query within the same source, scoring both semantic similarity (cosine similarity under an independent embedding model) and lexical similarity (token Jaccard and sequence ratio), and we check whether their gold sets overlap. The mean nearest-neighbor cosine similarity is 0.55--0.62 per dataset, reflecting shared in-domain topics rather than duplication. Combined leakage, defined as a synthetic query that is both a near-duplicate of an evaluation query (cosine similarity $\ge0.95$ or sequence ratio $\ge0.9$) and shares more than half of its gold set, occurs for 0 queries in five of the six settings and for 2 of 2{,}147 queries in Spider (0.09\%). Synthesis never accesses evaluation queries or relevance labels.

\section{Leave-One-Corpus-Out}
\label{app:loo}
Table~\ref{tab:loo} reports held-out performance when a corpus is removed entirely from synthesis (its documents appear as neither positives nor negatives). Performance on the held-out corpus remains close to that of the full model and far above that of the base embedder, indicating a transferable schema-retrieval ability.

\begin{table}[h]
\centering
\footnotesize
\setlength{\tabcolsep}{4pt}
\begin{tabular}{lccc}
\toprule
Held-out setting & Base & Full & Leave-out \\
\midrule
BEAVER (value)        & 12.5 & 48.4 & 44.3 \\
BEAVER (metadata)     & 42.2 & 55.4 & 57.7 \\
LiveSQLBench-Lg (tab) & 59.1 & 74.1 & 69.3 \\
LiveSQLBench-Lg (col) & 20.5 & 33.5 & 32.0 \\
\bottomrule
\end{tabular}
\caption{Leave-one-corpus-out recall@10 ($\times100$). \emph{Leave-out} never sees the held-out corpus during synthesis or training.}
\label{tab:loo}
\end{table}

\section{Prompt Templates}
\label{app:prompts}
We use two synthesis prompts, shown below. Both condition \emph{only} on schema documents drawn from the target corpus (field descriptions, column lists, and sample values that are part of the indexed documents); neither references any evaluation query or relevance label. The explicit instruction not to mention column or table names further discourages surface-form copying, complementing the leakage audit (Appendix~\ref{app:leakage}). Braces denote slots that are filled per specification.

\begin{tcolorbox}[breakable, colback=black!3, colframe=black!45,
  title=Column-level query synthesis, fonttitle=\bfseries\small,
  fontupper=\scriptsize]
\begin{Verbatim}[breaklines=true,breakanywhere=true,fontsize=\scriptsize]
You are a data analyst writing ONE realistic natural-language question
for the database "{DB}".

The question must be answerable ONLY by using ALL of these fields together
(it must genuinely require every one of them, not a subset):
{FIELD CARDS}   # one per gold column:
                #   `table.column` (type): description  e.g. v1, v2, v3
{JOIN LINE}     # join specs only:
  These fields live in DIFFERENT tables connected by foreign keys;
  the question must require combining them (a join).

{STYLE}         # one of:
  plain : Phrase it as a clear, direct business question a real analyst
          would ask.
  idiom : Use figurative / colloquial business language (e.g. 'pulling
          its weight', 'red flags', 'going against the grain'); it should
          still be a precise request underneath the figurative wording.
  kb    : Frame the question around this domain concept (ask for it in
          plain words, do NOT name the formula or its columns): {CONCEPT}
{FORBIDDEN}     # when hide-tokens are set:
  FORBIDDEN words: do NOT use these words or close variants in the
  question, describe what they mean instead: {TOKENS}

Hard rules:
- Write a real user request in natural language (1-2 sentences).
- Do NOT mention any column names, table names, SQL, or the database name.
- The question must require ALL the listed fields.

Output ONLY a JSON object: {"query": "<the question>"}
\end{Verbatim}
\end{tcolorbox}

\begin{tcolorbox}[breakable, colback=black!3, colframe=black!45,
  title=Table-level query synthesis, fonttitle=\bfseries\small,
  fontupper=\scriptsize]
\begin{Verbatim}[breaklines=true,breakanywhere=true,fontsize=\scriptsize]
You are a data analyst. Write ONE realistic natural-language question
whose answer requires the following data from database "{DB}":
{GROUNDING}     # single table: "a single table whose fields are: c1, c2, ..."
                # join: one line listing each connected table's fields
{NEED}          # single: "The question should be answerable from this one table."
                # join  : "The question MUST require combining data from ALL N
                #          tables (a multi-table join)."
{STYLE}         # short (BIRD/Spider): "Keep it SHORT and direct, about W words"
                # otherwise:           "About W words, a realistic analyst question"

Hard rules:
- A real user question in natural language.
- Do NOT mention table names, column names, SQL, or the database name.
- Ask about the real-world entities / data these tables hold.

Output ONLY a JSON object: {"query": "<the question>"}
\end{Verbatim}
\end{tcolorbox}

\end{document}